\documentclass[doublespacing,singlecolumn,11pt]{article}

\date{}

\usepackage{amsthm,amsmath,amssymb}
\usepackage{algorithm}
\usepackage{algorithmicx}
\usepackage{algpseudocode}
\usepackage{epsfig}
\usepackage{pifont}
\usepackage[utf8]{inputenc} 

\usepackage{float}

\usepackage{verbatim}
\usepackage{authblk}
\usepackage{booktabs}
\usepackage{multirow}
\usepackage{threeparttable}

\usepackage{geometry}
\usepackage[numbers]{natbib}

\newcommand\keywords[1]{\textbf{Keywords:} #1}
\newcommand{\etal}{\textit{et al}.}

\begin{document}

\title{PARF-Net: integrating pixel-wise adaptive receptive fields into hybrid Transformer-CNN network for medical image segmentation}
\author[1]{Xu Ma}
\author[1]{Mengsheng Chen}
\author[1]{Junhui Zhang}
\author[1]{Lijuan Song}
\author[1]{Fang Du}
\author[1]{Zhenhua Yu \thanks{Corresponding author: zhyu@nxu.edu.cn}}
\affil[1]{School of Information Engineering, Ningxia University}

\renewcommand*{\Affilfont}{\small\it}
\renewcommand\Authands{ and }

\maketitle

\begin{abstract}
	Convolutional neural networks (CNNs) excel in local feature extraction while Transformers are superior in processing global semantic information. By leveraging the strengths of both, hybrid Transformer-CNN networks have become the major architectures in medical image segmentation tasks. However, existing hybrid methods still suffer deficient learning of local semantic features due to the fixed receptive fields of convolutions, and also fall short in effectively integrating local and long-range dependencies. To address these issues, we develop a new method PARF-Net to integrate convolutions of Pixel-wise Adaptive Receptive Fields (Conv-PARF) into hybrid Network for medical image segmentation. The Conv-PARF is introduced to cope with inter-pixel semantic differences and dynamically adjust convolutional receptive fields for each pixel, thus providing distinguishable features to disentangle the lesions with varying shapes and scales from the background. The features derived from the Conv-PARF layers are further processed using hybrid Transformer-CNN blocks under a lightweight manner, to effectively capture local and long-range dependencies, thus boosting the segmentation performance. By assessing PARF-Net on four widely used medical image datasets including MoNuSeg, GlaS, DSB2018 and multi-organ Synapse, we showcase the advantages of our method over the state-of-the-arts. For instance, PARF-Net achieves 84.27\% mean Dice on the Synapse dataset, surpassing existing methods by a large margin.
\end{abstract}

\keywords{Medical image segmentation, Transformer, Convolutional neural network, Adaptive receptive fields}

\section{Introduction}
\label{sec:introduction}

Medical images contain a wealth of information about the internal structures of tissues, and are widely used in clinical disease diagnosis. As an auxiliary diagnostic tool, automatic medical image segmentation, such as segmentation of nuclei in histopathological slice images~\citep{zhang2024dawn} and lung CT segmentation~\citep{skourt2018lung}, can provide reliable assistance for medical professionals in pixel-wise localization and identification of the lesions. Accurate and robust pixel-level segmentation results will deliver a sound basis for computer-aided disease diagnosis and subsequent disease treatment. However, precisely disentangling lesions of varying shapes and scales from the background is still a challenging problem in medical image analysis.

As the predominant architecture in vision models~\citep{he2016deep,huang2017densely}, convolutional neural networks (CNNs) have shown superior ability of extracting semantically meaningful features from the natural images. Given the great successes of CNN models in vision tasks, many CNN-based medical image segmentation methods have been proposed in recent years~\citep{gu2019net,li2021applications,ronneberger2015unet,zhou2019unet++,isensee2021nnu}. These approaches generally employ a U-shaped network architecture with an encoder-decoder structure~\citep{ronneberger2015unet}. The encoder is comprised of multiple convolutional and down-sampling layers, responsible for progressively capturing multi-scale semantic features from the input medical image. The decoder gradually up-samples the compressed feature maps back to the original resolution, and yields the segmentation mask. Skip connections are built between the same-level encoding and decoding layers to restore the lost information of spatial details during down-sampling. U-Net~\citep{ronneberger2015unet} and its variants (e.g., UNet++~\citep{zhou2019unet++}, nnU-Net~\citep{isensee2021nnu} and MultiResUNet~\citep{ibtehaz2020multiresunet}) have achieved commendable segmentation results on various medical images from different modalities. Despite that CNN-based methods are excellent in capturing contextual information, the small convolutional kernels of CNNs make the network less effective in focusing the global features and modeling long-range dependency information, which are important for reliable segmentation of the lesions with varying shapes and scales in medical images. Another shortcoming lies in the fact that the weights of convolutional kernels are fixed after training, and the static kernels cannot well adapt to the complex contents of the input images.

Due to its high effectiveness in modeling long-range dependencies, the Transformer model~\citep{vaswani2017attention} built on self-attention mechanism has been an emerging backbone in the realm of natural language processing (NLP). To overcome inherent limitations of CNNs, Vision Transformer~\citep{dosovitskiy2020image} is the first work that introduces the self-attention mechanism into image classification tasks, and performs competitively with its CNN counterparts. The Transformers also yield the state-of-the-art performance in other vision tasks, such as semantic segmentation~\citep{strudel2021segmenter} and object detection~\citep{misra2021end}. Given its great success in natural image processing and analysis, the application of Transformers to medical images has attracted extensive attention in recent years~\citep{chen2021transunet, valanarasu2021medical,cao2022swin,Huang2023missformer}. For instance, TransUNet~\citep{chen2021transunet} is the first work that employs Transformer layers in encoder of the U-Net, to learn meaningful latent features from medical images; Medical Transformer~\citep{valanarasu2021medical} uses a gated axial-attention model to segment medical images; Swin-UNet~\citep{cao2022swin} is a pure Transformer with U-shaped network structure for medical image segmentation. These methods show strong performance in disentangling lesions from the noisy background. Despite that Transformers are excellent for modeling long-range dependencies, they are computational expensive due to quadratic time complexity, and deficient in capturing locality and translation invariance, leading to low inductive bias in representing local contextual information~\citep{xu2021vitae}.

As convolutions have property of translation invariance and locality, and Transformers excel in extracting global features and long-range dependencies, a natural idea is to integrate CNNs and Transformers into a hybrid architecture, which can enhance their advantages while alleviate the effects of their weaknesses. As a result, an increasing number of methods have been proposed to combine CNNs with Transformers for natural image analysis~\citep{zhang2023lite,liu2023learned,xia2024vit} and medical image segmentation~\citep{chen2021transunet, wang2022uctransnet,he2023h2former,lin2023lighter,lin2023convformer,yuan2023effective}. These methods exploit the mutually complementary roles of the two types of models to boost prediction accuracy. For instance, UCTransNet~\citep{wang2022uctransnet} replaces the skip connections with Transformer blocks to fuse multi-scale features derived from the encoder layers; H2Former~\citep{he2023h2former} adopts an encoder constituted by hierarchical CNN-Transformer blocks; CTC–Net~\citep{yuan2023effective} uses a feature complementary module to fuse features from a CNN encoder and a Transformer encoder. 

Although the existing hybrid methods exhibit strong performance across different image modalities, they still have several drawbacks that limit their applications. Firstly, most of the methods still suffer deficient learning of local and global features. For instance, CTC–Net~\citep{yuan2023effective} employs parallel CNN and Transformer encoders to fuse the same-level local and global features using a feature complementary module, while the two independent encoders involve in a large number of parameters and require extensive computation;  H2Former~\citep{he2023h2former} combines CNN and Transformer under a hierarchical structure, i.e. local features learned from a CNN module are fed into a Transformer block, which may limit the learning of long-range dependencies if local information is not well captured. Attempts have been made to find lightweight hybrid architectures by pruning~\citep{lin2023lighter} or designing CNN-style Transformers~\citep{lin2023convformer}. Secondly, existing hybrid methods often use fix-sized convolutional kernels to extract local features for all pixels, which may not well adapt to the complex contents of medical images, as the pixels located in the boundaries of the lesions with varying shapes and scales show high heterogeneity in semantics with the pixels of the background. Adaptive receptive fields have shown high effectiveness in segmentation or other vision tasks owing to their strong representation ability for different-sized objects~\citep{wei2017learning,jing2018stroke,Gao2020Deformable,xu2022arf}. However, hybrid architectures integrating Transformers and CNNs of adaptive receptive fields have not been yet fully investigated in existing medical image segmentation works. 

To address aforementioned issues, we propose a novel method called PARF-Net for medical image segmentation. In PARF-Net, Convolutions of Pixel-wise Adaptive Receptive Fields (Conv-PARF) are integrated into a hybrid Transformer-CNN U-Net to strengthen the feature representation ability. The Conv-PARF is used in the encoder to extract distinguishable features from input images, and is responsible for dynamically adjusting convolutional receptive fields for each pixel according to the preference of the pixel to a given kernel size. The preference is measured using the spatial attention mechanism previously proposed in~\citep{woo2018cbam}. Using the pixel-wise adaptive receptive fields is beneficial for extracting local contextual information from the organs or lesions with varying shapes and scales in medical images, thus improving the disentanglement of the targets from the background. Additionally, inspired by the approaches in~\citep{liu2021swin,yuan2023effective,liu2023learned}, we use hybrid Transformer-CNN layer consisting of two cascaded Transformer-CNN modules to capture local and global dependencies based on the features from the Conv-PARF modules. Each of the hybrid module combines a CNN block with window-based (shifted window-based) Transformer block under a parallel manner. To alleviate computational burden, input feature map of the Transformer-CNN module is equally divided into two parts, and the CNN (Transformer) block is responsible for processing one of the sub feature maps. The hybrid architecture effectively integrates the merits of CNN and Transformer, and simultaneously enables learning of local and global features under a lightweight and efficient manner. In summary, our contributions are three-fold:
\begin{itemize}
	\item Convolutions of pixel-wise adaptive receptive fields are proposed to learn distinguishable features from the medical image. The Conv-PARF copes with inter-pixel differences and dynamically adjusts convolutional kernels for each pixel, such that the organs or lesions with varying shapes and scales can be well captured and distinguished from the noisy background.
	\item We introduce hybrid Transformer-CNN modules into U-Net to efficiently extract and fuse global and local features. The hybrid architecture is implemented as a pair of parallel Transformer and CNN blocks, each of which is responsible for processing half of the input data, thus integrating the advantage of Transformer in modeling long-range dependencies and the merit of CNN in local feature learning.
	\item We comprehensively assess the performance of our method on four datasets, including MoNuSeg~\citep{kumar2017dataset,kumar2019multi}, GlaS~\citep{sirinukunwattana2017gland}, Data Science Bowl Challenge 2018 (DSB2018) and multi-organ Synapse~\citep{landman2015miccai}. By comparing PARF-Net with the state-of-the-arts (SOTAs) in terms of segmentation accuracies, we demonstrate that our method performs better than the existing methods.
\end{itemize}

\section{Related Work}
\subsection{CNN-based Models}
CNN initially achieves significant success in various fields of computer vision and is subsequently applied to medical image segmentation. The U-Net model~\citep{ronneberger2015unet} marks a milestone in this field. It is specifically designed to address the challenge of medical image segmentation and demonstrates notable performance, particularly when handling limited datasets. U-Net and its variants quickly become standard tools in the field of medical image segmentation. Skip connections in U-Net aid in the fusion of low-level details with high-level features, and enhance the integration of local features to improve the network's ability. For instance, UNet++~\citep{zhou2019unet++} is featured by nested dense skip connections with deep supervision, thus being able to extract multi-scale features effectively. Xiao~\etal~\citep{xiao2018weighted}~combine residual connections with U-Net to develop a new method called Res-UNet, where residual connections inserted between convolutional blocks assist in extracting irregular vascular information. Alom~\etal~\citep{alom2018recurrent}~introduce the Recurrent Residual Convolutional Neural Network (R2U-Net), in which recurrent convolutions serve as the backbone to facilitate feature accumulation. To address the lack of multi-scale feature analysis capability of U-Net and the semantic gap between encoder and decoder features, Ibtehaz~\etal~\citep{ibtehaz2020multiresunet}~propose MultiResUNet. Oktay~\etal~\citep{oktay2018attention}~introduce the Attention U-Net by integrating attention mechanisms with U-Net. They propose a novel attention gate mechanism, which highlights salient features during segmentation while suppressing irrelevant information. Jha~\etal~develop ResU-Net++~\citep{jha2019resunet++} and DoubleUNet~\citep{jha2020doubleu}, incorporating Squeeze \& Excitation blocks and atrous spatial pyramid pooling to expand the receptive field and capture high-resolution information. DCI-UNet~\citep{9976188} improves the U-Net architecture by introducing dilated convolution blocks and dilated inception blocks. U-Net along with its diverse variations usually employ a symmetric encoder-decoder structure and extract multiscale features of an image through convolutional operations. However, such network structures have limitations in capturing long-range dependencies in medical images, due to the limited receptive fields of convolutional kernels.

\subsection{Transformer-based Models}
Transformers are initially proposed in the field of NLP and achieve remarkable success owing to their superior ability of modeling global dependencies. Following their success in NLP, researchers introduce the pioneering Vision Transformer (ViT), which fundamentally transforms the landscape of image processing. Valanarasu~\etal~\citep{valanarasu2021medical}~propose a pure Transformer-based variant of U-Net in MedT. MedT introduces a gated axial-attention mechanism within the self-attention module, which restricts irrelevant information by adjusting the embedding positions of quary, key, and value. ViT has relatively high computational demands, particularly when processing high-resolution medical images. To reduce the computational burden and enhance model efficiency, Cao~\etal~\citep{cao2022swin}~adopt Swin-T blocks as the backbone of their network and proposes Swin-UNet. Swin-UNet divides input images into patches, which are sequentially fed into the encoder to learn both local and global features, leveraging a sliding window mechanism to extract contextual information. Despite that Swin-T achieves better computational efficiency compared to ViT, its model structure is complex and requires specific tuning expertise to achieve optimal performance. Huang~\etal~\citep{Huang2023missformer}~propose MISSFormer by redesigning the feed-forward network within the transformer blocks of the U-shaped encoder-decoder structure. This reconfiguration integrates local and global information, enhancing feature discrimination and thereby improving segmentation accuracy. Lin~\etal~\citep{10101800}~improve vanilla transformers by introducing cross-scale global transformer and boundary-aware local transformer modules.

\subsection{Hybrid Transformer-CNN based Models}
CNNs have strong image feature extraction capabilities and perform well in capturing local features, while Transformers can capture long-range dependencies, offering exceptional global modeling capabilities. Combining these two approaches preserves the strengths of both CNNs and Transformers, leading to further performance improvements. In recent years, hybrid Transformer-CNN architectures have attracted a lot of attention in medical image segmentation~\citep{he2023h2former,yuan2023effective,10506681,10520908}. Chen~\etal~\citep{chen2021transunet}~introduce TransUNet, the first approach to combine CNN and Transformer for medical image segmentation. This model can retain high-resolution spatial information while preserving low-resolution detail, leading to precise segmentation results. This demonstrates the capability of hybrid architectures to serve as an effective encoder for medical image segmentation. However, small-scale datasets often cannot meet the pretraining demands of Transformers, potentially reducing model performance. In addition, the fixed feature sequence size in ViT models limits local interaction information, which can result in loss of fine details, ultimately impacting segmentation efficiency. Yuan~\etal~\citep{yuan2023effective}~propose CTC-Net, which cleverly uses Swin-T to address the fixed feature sequence limitation of ViT models. CTC-Net employs parallel CNN and Transformer encoders to generate complementary features, which are effectively fused in a feature complementarity module. The use of two parallel and independent encoders make CTC-Net have a relatively large number of model parameters, making the training process much complex. Wang~\etal~\citep{wang2022uctransnet}~identify limitations in the original skip connections of U-Net, and propose UCTransNet, where Channel Transformer is introduced to replace the traditional skip connections for multi-scale feature analysis. This approach captures more complex dimensional relationships, bridging the semantic gap between encoder and decoder features, thereby enabling precise segmentation. He~\etal~\citep{he2023h2former} propose H2Former, a hierarchical hybrid model that integrates multi-scale channel attention (MSCA), CNN layers, and Transformer layers within a unified module. For lesion regions of varying sizes and shapes, MSCA can capture feature variations across different scales, thereby enhancing segmentation accuracy. HST-MRF~\citep{10520908} adopts a heterogeneous Swin Transformer that can fuse multi-resolution patch information for medical image segmentation. Unlike existing methods, our proposed PARF-Net combines convolutions of pixel-wise adaptive receptive fields with lightweight hybrid Transformer-CNN modules, and can effectively extract different-scale features from medical images.

\begin{figure}[htp]
	\centering
	\includegraphics[width=0.95\linewidth]{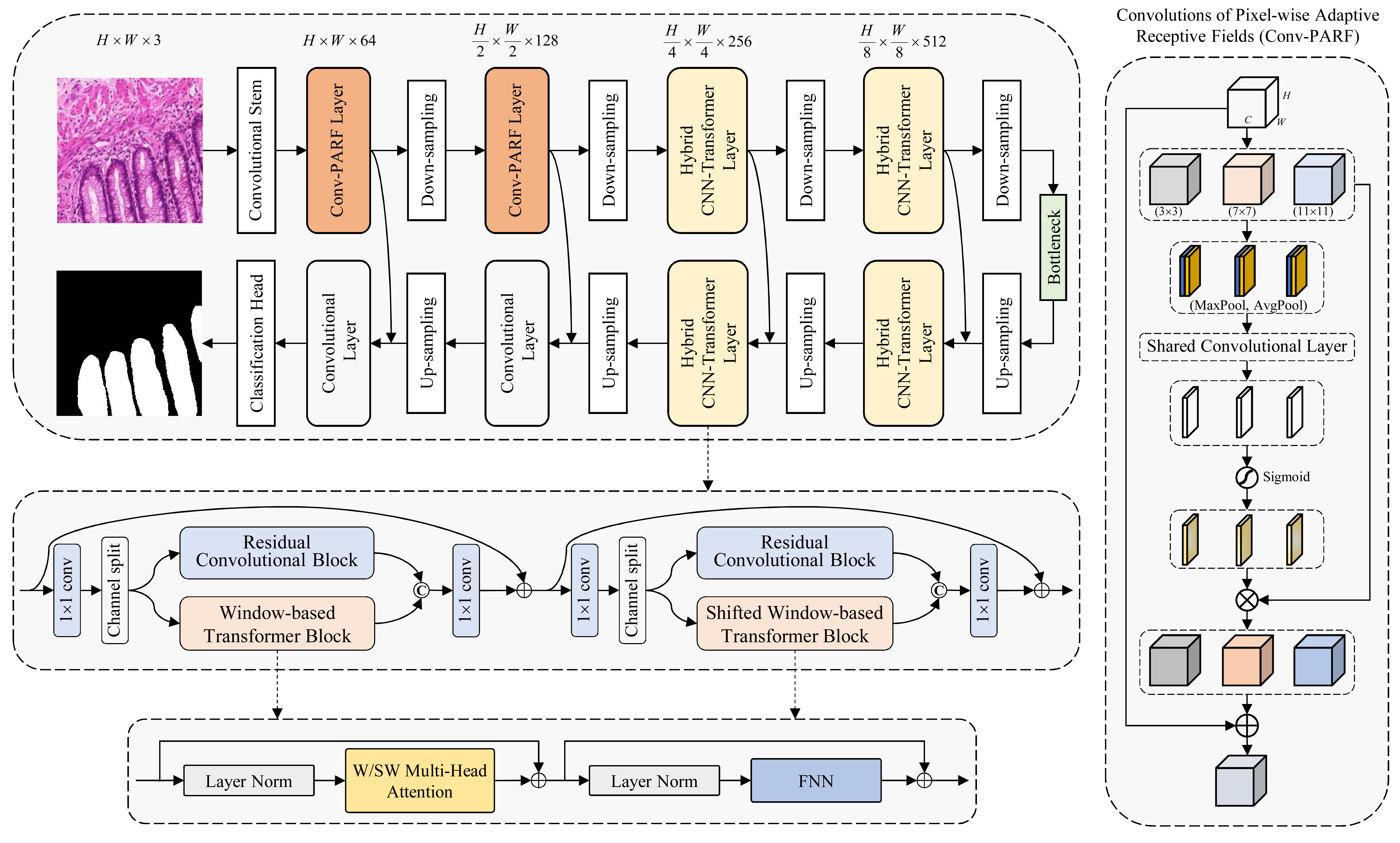}
	\caption{The model architecture of proposed PARF-Net. A U-shaped network that integrates convolutions of pixel-wise adaptive receptive fields (Conv-PARF) and hybrid Transformer-CNN modules is employed in PARF-Net. The Conv-PARF layers utilize spatial attention mechanism to dynamically extract spatial local information with the pixel-wise adaptive receptive fields. The hybrid modules are responsible for learning and fusing local and long-range dependencies, providing high-quality low-resolution features.} 
	\label{fig:fig1}
\end{figure}

\section{Methods}
In this paper, we introduce PARF-Net, a U-shaped Transformer-CNN network with convolutions of pixel-wise adaptive receptive fields for medical image segmentation. As shown in Figure~\ref{fig:fig1}, we enhance the ability of U-Net by introducing Conv-PARF in the encoder and inserting hybrid Transformer-CNN layers near the bottleneck. The Conv-PARF layer consists of a feature extraction module and a feature fusion module, the former is used to extract multi-scale features using convolutional kernels of different receptive fields, and the later is employed to integrate the multi-scale features using the activation maps corresponding to the receptive fields. We use Conv-PARF in the first two encoding layers to learn distinguishable local features from the input medical image, thus strengthening the quality of low-resolution features obtained from the subsequent encoding layers, and providing a sound basis for the classification head to generate accurate segmentation mask. Two hybrid Transformer-CNN layers are placed before and after the bottleneck to efficiently fuse global and local features. On the one hand, the hybrid modules are more suitable for mining the high-level features containing meaningful semantic information of the organs or lesions. On the other hand, employing the hybrid modules to process the low-resolution features is computationally efficient and delivers rich information to gradually restore the high-resolution features. The following subsections provide a detailed description of the model components of PARF-Net.

\subsection{Convolutions of pixel-wise adaptive receptive fields}
Adaptive receptive fields settle the limitations of static convolutions in capturing geometric variations of the objects in an image, and have enabled improved performance in various vision tasks~\citep{wei2017learning, jing2018stroke, Gao2020Deformable, xu2022arf}. Wei~\etal~\citep{wei2017learning} propose to implicitly change receptive field of a layer by inflating/shrinking the feature maps of its precursor layer. Jing~\etal~\citep{jing2018stroke} employ adaptive receptive fields to achieve stroke controllable style transfer. To handle object deformations, Gao~\etal~\citep{Gao2020Deformable} propose deformable kernels to adapt the effective receptive field. For breast mass segmentation, Xu~\etal~\citep{xu2022arf} propose to select suitable receptive fields according to the size of the object. Ma~\etal~\citep{ma2021improving} introduce adaptive receptive fields in graph neural networks to improve the performance of node representation learning by optimizing receptive field for each node. Given these successful applications of adaptive receptive fields, we propose pixel-wise adaptive receptive fields for medical image segmentation.

Given the input feature map $x \in \mathbb{R}^{H\times W\times C}$ ($H$, $W$ and $C$ denote the height, width and number of channels of the feature map, respectively), the Conv-PARF layer first uses a set of predefined convolutional kernels (e.g., $3 \times 3$, $7 \times 7$, $11 \times 11$) to extract features under different receptive fields, resulting a set of feature map $\mathcal{F}=\{\mathcal{F}_k \vert \mathcal{F}_k \in \mathbb{R}^{H\times W\times C},k = 1,2,\ldots,K\}$, here $K$ denotes the number of receptive fields. Then, the spatial attention is applied to each $\mathcal{F}_k$ to get the preference score of each pixel to the $k$-th receptive field. Concretely, the maximum and average values of $\mathcal{F}_k$ along the channel dimension are calculated for each pixel, giving a feature map with size of $H\times W\times 2$, then a shared $7\times 7$ convolutional layer and the Sigmoid activation function are leveraged to get the activation map $A_k \in \mathbb{P}^{H\times W }$ corresponding to the $k$-th receptive field. With the activation maps of all receptive fields, we get the resulted feature map $y \in \mathbb{R}^{H\times W\times C}$ as follows:
\begin{equation}
	y = x + \sum_{k=1}^K A_k*\mathcal{F}_k
\end{equation}
where $A_k$ acts as the weights of the $k$-th feature map $\mathcal{F}_k$, and a skip connection is built between the input $x$ and the output. The activation map $A_k$ is responsible for switching up or off the $k$-th receptive field for each pixel, and fusing the features $\mathcal{F}$ by the activation maps delivers the convolutional results under pixel-wise adaptive receptive fields.

\subsection{Hybrid Transformer-CNN modules}
We use hybrid Transformer-CNN layers to fuse global and local features, by capitalizing on CNN’s local feature extraction strengths and the Transformer’s ability to model global dependencies. The Transformer is responsible for modeling long-range dependencies from the image, while the CNN focuses on extracting finer local features. The introduced Transformer-CNN layer is comprised of two cascaded Transformer-CNN modules, each of which contains a pair of CNN and window-based Transformer (or shifted window-based Transformer) blocks. Given the input feature map $x \in \mathbb{R}^{H\times W\times C}$, the workflow of the Transformer-CNN layer includes: 1) convolutions of 1$\times$1 kernels are used to fuse the features along the channel dimension, then a split operation is employed to divide the feature map into two equal-sized parts $x_c \in \mathbb{R}^{H\times W\times \frac{C}{2}}$ and $x_t \in \mathbb{R}^{H\times W\times \frac{C}{2}}$; 2) the feature map $x_c$ is processed using a residual convolutional block, formed by two convolutional layers with 3$\times$3 kernels and the LeakyReLU activation function, to capture local features $x_{cl} \in \mathbb{R}^{H\times W\times \frac{C}{2}}$, while the feature map $x_t$ is fed into a window-based Transformer block, which utilizes window-based multi-head self-attention (W-MSA) mechanism to capture global dependencies, generating feature map $x_{tg} \in \mathbb{R}^{H\times W\times \frac{C}{2}}$; 3) $x_{cl}$ and $x_{tg}$ are concatenated along the channel dimension and then processed using convolutions of 1$\times$1 kernels, which allows the model to exploit both global semantics and local details when performing feature extraction; 4) a skip connection is constructed between the input $x$ and the output to allow for more efficient feature fusion, which gives the final feature map $y_w \in \mathbb{R}^{H\times W\times C}$ of the first Transformer-CNN module; and 5) the feature map $y_w$ is handled with the second Transformer-CNN module, where the shifted window-based multi-head self-attention (SW-MSA) mechanism is used to learn non-local features, while other components are unchanged. The output feature map of the Transformer-CNN layer is denoted as $y \in \mathbb{R}^{H\times W\times C}$. In summary, the pipeline of the first Transformer-CNN module can be formulated as follows:
\begin{equation}
	x_c, x_t = \text{Split}\left  (\text{Conv}_{1 \times 1}(x) \right)
\end{equation}

\begin{equation}
	x_{cl}, x_{tg} = \text{RC}(x_c), \text{WT}(x_t)
\end{equation}

\begin{equation}
	y_w = x + \text{Conv}_{1 \times 1}\left( \text{Cat}(x_{cl}, x_{tg}) \right)
\end{equation}
where RC and WT represent the residual convolution and window-based Transformer, respectively. Similarly, the pipeline of the second Transformer-CNN module can be defined as follows:
\begin{equation}
	y_c, y_t = \text{Split}\left  (\text{Conv}_{1 \times 1} (y_w) \right)
\end{equation}

\begin{equation}
	y_{cl}, y_{tg} = \text{RC}(y_c), \text{SWT}(y_t)
\end{equation}

\begin{equation}
	y = y_w + \text{Conv}_{1 \times 1}\left( \text{Cat}(y_{cl}, y_{tg}) \right)
\end{equation}
where SWT denotes the shifted window-based Transformer. The hybrid Transformer-CNN modules exploit the advantage of Swin Transformer in efficiently modeling global dependencies, and subtly combine it with the residual convolutions under a lightweight manner, which promises both effectiveness and efficiency in feature learning.

\subsection{Overall architecture of PARF-Net}
Inspired by the success of U-Net and its variants in medical image segmentation, our proposed PARF-Net is a U-shape architecture consists of four encoding layers and four corresponding decoding layers. As the original medical image is of high pixel-to-pixel variance, it is more appropriate to use the Conv-PARF instead of conventional static convolutions for extracting features for the pixels. Therefore, we introduce Conv-PARF to the first two encoding layers of PARF-Net to abstract local contextual information from input medical image. Specifically, the channel dimension of input medical image is expanded to 64 using a convolutional stem, then the features are processed with two Conv-PARF layers, each of which is followed by a down-sampling module that reduces the size of the feature map into half while doubles the channel dimension. The informative features derived from the Conv-PARF layers are beneficial for acquiring high-quality low-resolution features via the subsequent encoding layers, and yielding precise segmentation mask through the classification head. The last two encoding layers are the Transformer-CNN layers accompanied with down-sampling modules, which shows two advantages: 1) the input feature maps of the Transformer-CNN layers are of low-resolution, thus largely mitigating the computational burden of the Transformer blocks; and 2) the hybrid modules can more easily capture the local and non-local patterns from the distinguishable features derived from the precursor Conv-PARF layers, thus boosting the performance of medical image segmentation. 

The decoder of PARF-Net is formed by two Transformer-CNN layers and two static convolutional layers. An up-sampling module is placed in front of each decoding layer to incrementally restore the feature resolution, and the resulted features are fused with these from the corresponding encoding layer, which allows for better recovery of details and contextual information. By exploiting the enhanced ability in capturing local and long-range dependencies, the Transformer-CNN layers utilize the low-resolution features from the encoder to generate high-resolution informative features, which will provide a sound basis for the subsequent convolutional layers to learn distinguishable features, enabling the classification head to produce accurate mask for segmentation.

\subsection{Model optimization}
Suppose $x_i$ denotes the input medical image, $y_i$ represents the corresponding segmentation mask, and $\hat{y}_i$ is the segmentation result of the model, we use a combination of cross-entropy loss and Dice loss to define the loss function:
\begin{equation}
	\mathcal{L} = \mathcal{L}_{ce}(y_i, \hat{y}_i) + \mathcal{L}_{dice}(y_i, \hat{y}_i)
\end{equation}
The model parameters are updated with the Adam optimization algorithm.

\subsection{Datasets}
\subsubsection{MoNuSeg dataset~\citep{kumar2017dataset,kumar2019multi}}
The MoNuSeg dataset is composed of 44 tissue images stained with Hematoxylin and eosin (H\&E), captured at a 40$\times$ magnification level. This collection includes tissues from seven different organ types, with both benign and malignant samples across various patients. It is split into 30 training images and 14 testing images.
\subsubsection{GlaS dataset~\citep{sirinukunwattana2017gland}}
The GlaS dataset consists of microtomes stained with H\&E and labeled by an experienced pathologist. 85 out of the 165 images are used for training and the remaining 80 images for testing.
\subsubsection{DSB2018 dataset}
The DSB2018 cell nucleus dataset is available at the Data Science Bowl Challenge 2018, which contains a total of 670 images with accurate annotations. We use 80\% images for training and the remaining 20\% for testing.
\subsubsection{Synapse dataset~\citep{landman2015miccai}}
The Synapse dataset is a benchmark for multi-organ segmentation research in medical image analysis. It comprises 30 abdominal organ CT scans (a total of 3779 axial clinical CT images), covering 8 organ classes: aorta, gallbladder, spleen, left kidney, right kidney, liver, pancreas and stomach. Following TransUNet~\citep{chen2021transunet}, 18 cases are used as the training set, and 12 cases are designated as the test set.

For all datasets, the images and corresponding masks are resized to 224$\times$224 for training and testing.

\subsection{Implementation details}
PARF-Net is implemented with PyTorch and trained on a computational server with two NVIDIA TITAN RTX graphics cards. The Adam optimization algorithm is employed to update the network weights of PARF-Net. When training all methods on GlaS, MoNuSeg and DSB2018 datasets, we apply horizontal and vertical flips as well as random rotations to increase the variability of training data, and set batch size, maximum training epoch and starting learning rate to 4, 400 and 0.0001, respectively. For the Synapse dataset, we set batch size to 4, and configure other hyper-parameters by following~\citep{Huang2023missformer}.

\subsection{Performance evaluation}
\subsubsection{Evaluation metrics}
On the datasets MoNuSeg, GlaS and DSB2018, we use Intersection-over-Union (IoU) and Dice Similarity Coefficient (Dice) as the main performance indicators, and additionally report pixel-wise Accuracy (ACC), Recall and Precision.

On the Synapse dataset, the evaluation metrics include the mean Dice and the mean Hausdorff Distance (HD) calculated from 2211 2D slices extracted from 3D volumes. Both Dice and HD are employed to assess segmentation performance for the eight abdominal organs.

\subsubsection{Comparison with other methods}
To assess the segmentation performance of our proposed PARF-Net model, we make comparisons between PARF-Net and the SOTAs. Three types of methods are used as the competitors: CNN based (V-Net~\citep{milletari2016v}, U-Net~\citep{ronneberger2015unet} and its variants), Transformer based (Swin-UNet~\citep{cao2022swin} and MISSFormer~\citep{Huang2023missformer}) and hybrid architecture based (TransUNet~\citep{chen2021transunet}, UCTransNet~\citep{wang2022uctransnet}, CTC-Net~\citep{yuan2023effective} and H2Former~\citep{he2023h2former}). We run each method five times on the datasets MoNuSeg, GlaS and DSB2018, and report the mean and standard variance of each performance metric. As model training on the Synapse dataset is time-consuming, we only report one experimental result for each method.

\begin{table}[htb]	
	\caption{Performance comparison results on the MoNuSeg dataset. Each method is run five times to obtain the mean and standard variance of each metric.}
	\label{tab:table1}
	\begin{center}
		\begin{tabular}{c|c|ccccc}
			\hline
			Type & Method & IoU$\uparrow$ & Dice$\uparrow$ & ACC$\uparrow$ & Recall$\uparrow$ & Precision$\uparrow$ \\
			\hline
			\multirow{8}{*}{CNN}
			&           V-Net~\citep{milletari2016v}          		& 55.42±1.62 & 71.30±1.35  & 84.87±0.95 & 85.01±2.69  &61.46±2.07  \\
			&           UNet~\citep{ronneberger2015unet}           	    & 62.98±0.23 & 77.25±0.18 & 89.60±0.23  & 78.64±0.76  &76.09±1.06  \\
			&           UNet++~\citep{zhou2019unet++}            	& 62.74±0.09 & 77.09±0.06 & 88.57±0.22 & 84.83±1.47  &70.75±1.08  \\
			&           Attention U-Net~\citep{oktay2018attention}     & 64.45±1.61 & 78.37±1.19 & 89.77±0.45 & 83.40±2.11  &74.15±1.2   \\
			&           ResU-Net++~\citep{jha2019resunet++}          & 61.06±1.24 & 75.82±0.96 & 85.83±5.59 & 82.93±1.93  &69.94±2.99  \\
			&           DoubleUNet~\citep{jha2020doubleu}          & 62.12±0.65 & 76.64±0.49 & 88.68±0.37 & \textbf{89.96±1.04}  &70.51±1.24  \\
			&           R2U-Net~\citep{alom2018recurrent}             & 63.18±0.54 & 77.40±0.41  & 89.63±0.22 & 78.60±1.2    &\textbf{76.50±1.1} \\
			&           MultiResUNet~\citep{ibtehaz2020multiresunet}        & 60.72±0.73 & 75.54±0.57 & 88.75±0.27 & 77.40±0.88   &73.94±0.79  \\
			
			\hline
			\multirow{2}{*}{Trans} 
			&           Swin-UNet~\citep{cao2022swin}           & 62.07±0.94 & 76±0.84    & 88.21±0.85 & 88.44±1.47  &70.67±2.09  \\
			&           MISSFormer~\citep{Huang2023missformer}          & 61.37±2.36 & 75.86±1.95 & 87.75±1.58 & 86.13±2.23  &67.8±3.63   \\
			
			\hline
			\multirow{3}{*}{Hybrid}
			&           TransUNet~\citep{chen2021transunet}          & 61.94±0.98 & 76.42±0.76 & 88.78±0.52 & 83.56±1.57  &71.2±1.8    \\
			&           UCTransNet~\citep{wang2022uctransnet}          & 62.36±0.83 & 76.82±0.63 & 88.96±0.51 & 82.48±1.78  &72.1±2      \\ 
			&           CTC-Net~\citep{yuan2023effective}             & 60.77±0.31 & 75.59±0.24 & 88.47±0.24 & 8052±0.82   &71.48±0.95      \\
			&           H2Former~\citep{he2023h2former}             & 57.25±0.46 & 72.83±0.37 & 86.56±0.56 & 80.99±2.77  &66.46±2.21  \\
			&           PARF-Net(Ours)       & \textbf{64.83±0.4} & \textbf{78.66±0.297} & \textbf{89.94±0.13} & 83.17±0.54  &74.71±0.13   \\
			\hline
		\end{tabular} 
	\end{center}
\end{table}

\begin{table}[htb]	
	\caption{Performance comparison results on the GlaS dataset. Each method is run five times to obtain the mean and standard variance of each metric.}
	\label{tab:table2}
	\begin{center}
		\begin{tabular}{c|c|ccccc}
			\hline
			Type & Method & IoU$\uparrow$ & Dice$\uparrow$ & ACC$\uparrow$ & Recall$\uparrow$ & Precision$\uparrow$ \\
			\hline
			\multirow{8}{*}{CNN} 
			&           V-Net~\citep{milletari2016v}         		& 80.45±0.42 & 89.16±0.26 & 88.72±0.33 & 91.3±0.63   &87.16±0.26  \\
			&           UNet~\citep{ronneberger2015unet}           	    & 81.06±1.48 & 89.47±0.86 & 88.91±1.03 & 90.85±0.68  &88.36±2.09  \\
			&           UNet++~\citep{zhou2019unet++}           	& 81.03±0.77 & 89.34±0.45 & 89.18±0.4  & 90.79±1.39  &88.19±0.62  \\
			&           Attention U-Net~\citep{oktay2018attention}     & 81.98±0.45 & 90.1±0.28  & 89.71±0.27 & 91.28±1.32  &87.88±3.09  \\
			&           ResU-Net++~\citep{jha2019resunet++}          & 79.5±0.97  & 88.57±0.6  & 88.01±0.84 & 90.6±1.04   &86.71±2.04  \\
			&           DoubleUNet~\citep{jha2020doubleu}          & 80.3±0.19  & 89.02±0.13 & 88.32±0.28 & 92.28±1.37  &86.09±1.3   \\
			&           R2U-Net~\citep{alom2018recurrent}             & 78.28±1.28 & 87.87±0.82 & 86.84±1.14 & 91.16±1.76  &84.88±2.68  \\
			&           MultiResUNet~\citep{ibtehaz2020multiresunet}        & 77.48±0.72 & 87.18±0.48 & 86.53±0.81 & 88.88±2.1   &85.99±2.19  \\
			\hline
			\multirow{2}{*}{Trans} 
			&           Swin-UNet~\citep{cao2022swin}           & 81.63±2.13 & 89.55±1.4  & 89.5±1.31  & 91.29±0.98  &88.55±2.47  \\
			&           MISSFormer~\citep{Huang2023missformer}          & 76.78±0.75 & 86.06±0.52 & 86.21±0.49 & 88.34±1.16  &85.9±1.04   \\
			\hline
			\multirow{3}{*}{Hybrid}
			&           TransUNet~\citep{chen2021transunet}           & 82.71±0.64 & 90.44±0.38 & 90.16±0.39 & 90.89±0.78  &\textbf{90.14±0.81}  \\
			&           UCTransNet~\citep{wang2022uctransnet}          & 81.17±0.71 & 89.45±0.45 & 89.22±.37  & 90.15±1.32  &89.04±0.5   \\ 
			&           CTC-Net~\citep{yuan2023effective}             & 83.14±0.69 & 88.83±3.05 & \textbf{90.72±0.53} & 92.72±0.28  &88.93±0.89  \\
			&           H2Former~\citep{he2023h2former}            & 82.78±0.7  & 90.47±0.43 & 90.12±0.39 & 91.14±1.06  &90.02±0.83  \\
			&           PARF-Net(Ours)      & \textbf{83.76±0.65} & \textbf{91.06±0.4}  & 90.67±0.34 & \textbf{92.9±1.23}   &89.53±0.72  \\
			\hline
		\end{tabular}    
	\end{center}
\end{table}

\section{Results and discussions}

\subsection{Results on MoNuSeg}
The quantitative results on the MoNuSeg dataset are shown in Table~\ref{tab:table1}. In general, our method outperforms the counterparts by exhibiting higher IoU, Dice and ACC. For instance, PARF-Net achieves mean IoU of 64.83\% and mean Dice of 78.66\%, surpassing the second-best method Attention U-Net by 0.38\% and 0.29\%, respectively. Additionally, our method shows significant advantage when compared to the hybrid counterparts, especially in IoU, Dice and Precision metrics. For instance, TransUNet reaches mean IoU of 61.94\% and mean Dice of 76.42\%, UCTransNet yields mean IoU of 62.36\% and mean Dice of 76.82\%, falling evidently short of the performance of PARF-Net. It is also noted that the pure Transformer and hybrid counterparts have little advantage over the pure CNN methods on this dataset, as high complexity of Transformer and hybrid models make them easily suffer over-fitting when trained on a small number of samples. By comparison, our method shows better generalization ability and well tackles such limitation, which benefits from the joint usage of the Conv-PARF that adaptively extracts spatial local information for each pixel, and the hybrid Transformer-CNN architecture that effectively processes the local features and long-range dependencies. The results imply effective learning of high-resolution features and fusion of same-level local and global features in PARF-Net have advantages over the SOTAs, and are more friendly to the segmentation task with limited training samples.

\subsection{Results on GlaS}
We then evaluate our method on the GlaS dataset, and comparison results with other methods are shown in Table~\ref{tab:table2}. Compared to the pure CNN or Transformer methods, the hybrid counterparts generally depict higher segmentation accuracies partly owing to the increased number of training images. For instance, TransUNet and CTC-Net achieve the highest Precision and ACC scores, respectively, while our method surpasses all the counterparts in the remaining three metrics by giving 83.76\% mean IoU, 91.06\% mean Dice and 92.9\% mean Recall. The superior performance of hybrid methods is built on the capability of exploiting both spatial local information and global dependencies to learn semantic features. Among the hybrid competitors, H2Former shows higher Dice due to the delicately designed hierarchical structure for integrating the merits of CNN and Transformer, but the quality of extracted spatial local features may affect the subsequent learning of long-range dependencies, and vice versa. By comparison, our method employs the Conv-PARF layers to extract local information, then independently processes local and global features under a parallel manner, alleviating the mutual adverse effects while supplementing each other. These characteristics enable improved segmentation performance of PARF-Net.

\begin{table}[htb]	
	\caption{Performance comparison results on the DSB2018 dataset. Each method is run five times to obtain the mean and standard variance of each metric.}
	\label{tab:table3}
	\begin{center}
		\begin{tabular}{c|c|ccccc}
			\hline
			Type & Method & IoU$\uparrow$ & Dice$\uparrow$ & ACC$\uparrow$ & Recall$\uparrow$ & Precision$\uparrow$\\
			\hline
			\multirow{9}{*}{CNN} 
			&           V-Net~\citep{milletari2016v}          		 & 86.92±0.16  & 92.89±0.09 & 97.17±0.03 & 93.71±0.35  &92.36±0.27 \\
			&           UNet~\citep{ronneberger2015unet}           	     & 85.52±0.37  & 92.16±0.21 & 96.8±0.1   & 92.73±0.62  &91.7±0.47  \\
			&           UNet++~\citep{zhou2019unet++}           	 & 87.08±0.06  & 93.08±0.04 & 97.22±0.01 & 93.42±0.46  &92.54±0.39 \\
			&           Attention U-Net~\citep{oktay2018attention}      & 86.6±0.23   & 92.73±0.14 & 97.29±0.06 & 92.73±0.67  &92.96±0.76 \\
			&           ResU-Net++~\citep{jha2019resunet++}           & 86.51±0.24  & 92.72±0.12 & 97.04±0.09 & 93.62±0.39  &91.95±0.42  \\
			&           DoubleUNet~\citep{jha2020doubleu}           & 86.05±0.16  & 92.46±0.09 & 96.9±0.05  & 93.6±0.52   &91.45±0.63\\
			&           R2U-Net~\citep{alom2018recurrent}              & 83.33±1.62  & 90.86±0.98 & 96.45±0.35 & 89.22±1.75  &92.75±0.87 \\
			&           MultiResUNet~\citep{ibtehaz2020multiresunet}         & 84.97±0.68  & 91.83±0.38 & 96.67±0.2  & 93.04±0.5   &90.75±0.59 \\
			
			\hline
			\multirow{3}{*}{Trans} 	
			&           Swin-UNet~\citep{cao2022swin}            & 87.9±0.1   & 93.5±0.06   & 97.52±0.02 & 93.17±0.25  &93.99±0.27 \\
			&           MISSFormer~\citep{Huang2023missformer}           & 86.24±0.28 & 92.53±0.16  & 97.18±0.06 & \textbf{96.24±0.34}  &92.63±0.46 \\
			
			\hline
			\multirow{4}{*}{Hybrid}
			&           TransUNet~\citep{chen2021transunet}            & 87.93±0.21 & 93.51±0.12  & 97.55±0.06 & 93.22±0.38  &94.04±0.57 \\
			&           UCTransNet~\citep{wang2022uctransnet}           & 87.72±0.2  & 93.41±0.14  & 97.5±0.02  & 92.97±0.27  &94±0.16   \\
			&           CTC-Net~\citep{yuan2023effective}              & 88.06±0.13 & 93.59±0.07  & 97.56±0.04 & 93.4±0.59   &93.89±0.62  \\
			&           H2Former~\citep{he2023h2former}             & 84.84±0.32 & 91.72±0.19  & 96.83±0.08 & 92.23±0.64  &91.46±0.61  \\
			&           PARF-Net(Ours)       & \textbf{88.97±0.27} & \textbf{94.14±0.18}  & \textbf{97.75±0.04} & 93.57±0.73  &\textbf{94.79±0.57} \\
			\hline
		\end{tabular}
	\end{center}
\end{table}

\begin{table}[htb]	
	\scriptsize
	\caption{Performance comparison results on the Synapse dataset.}
	\label{tab:table4}
	\begin{center}
		\begin{tabular}{c|c|cc|cccccccc}
			\hline
			Type& Method & Dice$\uparrow$ & HD$\downarrow$ & Aorta & Gallbladdr & Kidney(L) & Kidney(R) & Liver & Pancreas & Spleen & Stomach \\
			\hline
			\multirow{2}{*}{CNN} 
			&           UNet~\citep{ronneberger2015unet}           	    & 82.76 & 26.55 & 89.69 &72.06 & 85.22 & 80.92 & 95.59 & 68.54 & 87.75 &78.91  \\ 
			&           Attention U-Net~\citep{oktay2018attention}     & 81.87 & 21.16 & 88.51 &68.58 & 79.18 & 80.54 & 94.51 & 67.51 & 89.06 &\textbf{82.87}  \\	
			\hline
			\multirow{2}{*}{Trans} 
			&           Swin-UNet~\citep{cao2022swin}           & 76.28 & 25.79 & 85    &55.5  & 77.61 & 74.59 & 93.33 & 58.48 & 89.45 &76.24  \\
			&           MISSFormer~\citep{Huang2023missformer}          & 79.15 & 28.33 & 86.09 &68.07 & 81.53 & 77.86 & \textbf{96.68} & 58.06 & 90.34 &77.35  \\
			
			\hline
			\multirow{4}{*}{Hybrid}
			&           TransUNet~\citep{chen2021transunet}           & 76.75 & 34.2  & 86.53 &65.08 & 79.35 & 73.96 & 93.91 & 54.32 & 85.86 &75.01  \\
			&           UCTransNet~\citep{wang2022uctransnet}          & 81.69 & 23.69 & 86.68 &62.61 & 85.62 & 82.89 & 94.45 & 66.97 & 90.66 &82.69  \\
			&           CTC-Net~\citep{yuan2023effective}             & 78.81 & 25.77 & 86.81 &60.24 & 84.67 & 83.13 & 94.12 & 54.46 & \textbf{91.8}  &75.15  \\
			&           H2Former~\citep{he2023h2former}            & 82.27 & 13.87 & 87.4  &68.1   &  \textbf{86.76} &  \textbf{84.1}  & 95.1  &67.1  &90.3 & 78.49\\ 
			&           PARF-Net(Ours)   	& \textbf{84.27} & \textbf{13.82} & \textbf{90.38} &\textbf{72.95} & 85.91 & 83.42 & 95.54 & \textbf{72.86} & 91.43 &78.91  \\
			\hline
		\end{tabular}
		
	\end{center}
\end{table}

\subsection{Results on DSB2018}
We further examine the segmentation performance of PARF-Net on the DSB2018 dataset, and the results are shown in Table~\ref{tab:table3}. This dataset contains 536 cell nucleus images for training and 134 images for testing. Our method outperforms other methods by achieving higher IoU, Dice, ACC and Precision scores. For instance, PARF-Net gets mean IoU of 88.97\% and mean Dice of 94.14\%, surpassing the second-best method CTC-Net by 0.91\% and 0.55\%, respectively. Similar to the results on the dataset GlaS, the pure CNN or Transformer methods generally fall short of the hybrid methods. The best CNN method UNet++ shows mean IoU of 87.08\%, and the best Transformer method Swin-UNet exhibits mean IoU of 87.9\%, while the hybrid method TransUNet beats them by yielding mean IoU of 87.93\%. These results further demonstrate that hybrid architectures have advantages in segmentation tasks with sufficient training medical images, owing to their higher model complexity and ability of capturing local details and long-range dependencies. Among the hybrid methods, our method again gets the best segmentation results, benefitting from the joint usage of Conv-PARF and hybrid architecture. 

\subsection{Results on Synapse}
Table~\ref{tab:table4} depicts the segmentation results of our method and eight SOTA counterparts on the 3D Synapse dataset. Two performance metrics, i.e. mean Dice and HD, are used to evaluate the methods, and we also report the Dice for each organ. The experimental results clearly showcase the advantage of our method over the counterparts in segmentation of multi-organ images. The proposed PARF-Net achieves the highest mean Dice of 84.27\% and lowest mean HD of 13.82 among all the methods. Additionally, our method outperforms the SOTAs on three categories, including Aorta, Gallbladdr and Pancreas. Particularly, all existing methods fall evidently short of PARF-Net on pancreas segmentation owing to the large deformation and blurred boundary of the pancreas. Our method achieves 72.86\% Dice on pancreas segmentation, exceeding the second-best UNet (Dice of 68.54\%) by a large margin. Thanks to the Conv-PARF that is able to tackle inter-pixel differences and adaptively extract spatial local information from the input image, and the Transformer-CNN architecture for effectively integrating local information and global interactions, our method generates the best results on segmentation of three out of eight organs among all the methods. Considering the counterparts have produced SOTA results, the better performance achieved by our proposed PARF-Net validates its superior ability of processing complex 3D images with large inter-organ differences.

\subsection{Visualization results}

\begin{figure}[htb]
	\centering
	\includegraphics[width=0.9\linewidth]{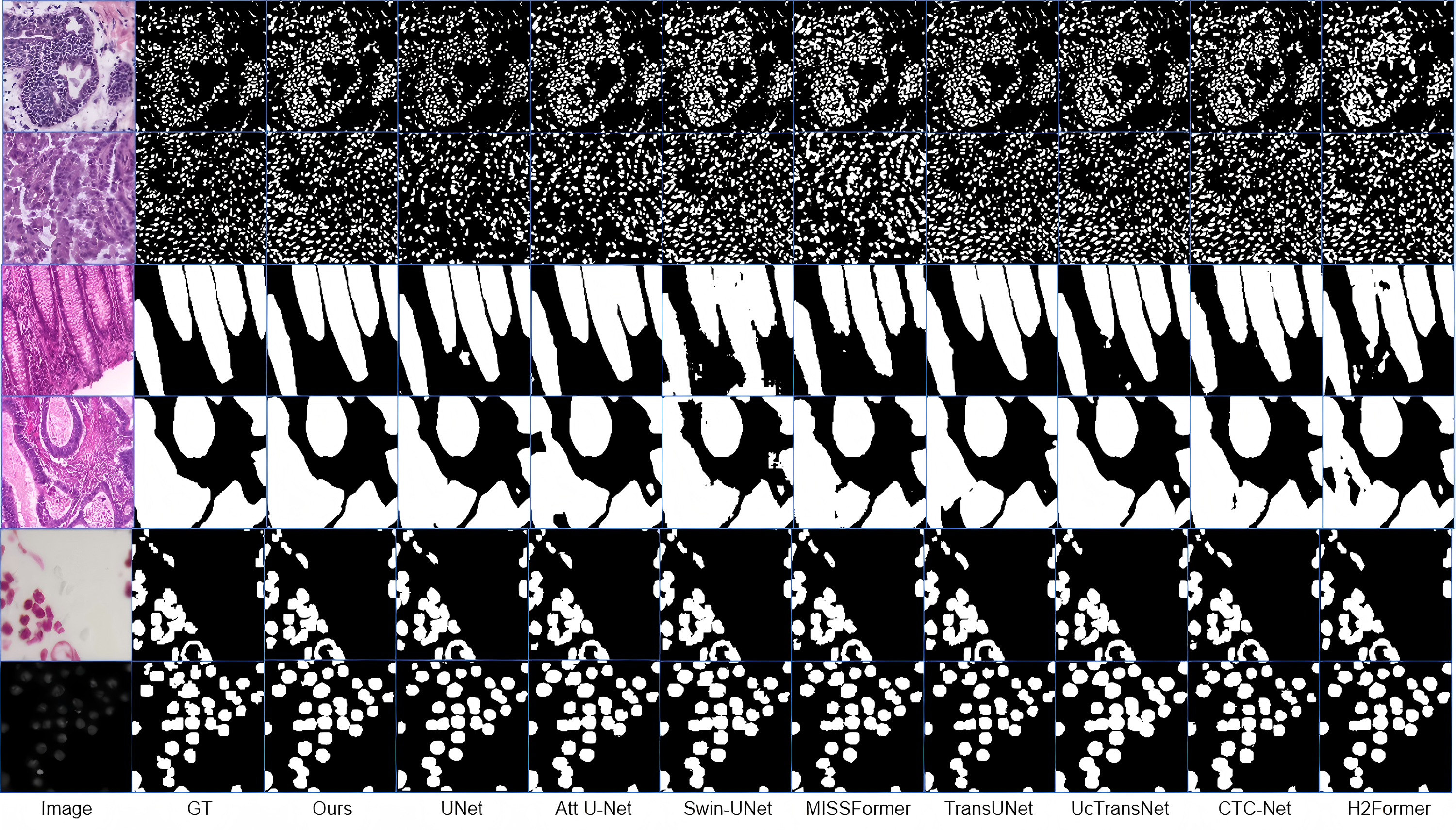}
	\caption{Visualization of segmentation results on the MoNuSeg, GlaS and DSB2018 datasets. For each dataset, two medical images accompanied with the ground truth segmentation masks and predicted results are shown.} 
	\label{fig:fig2}
\end{figure}

\begin{figure}[htb]
	\centering
	\includegraphics[width=0.9\linewidth]{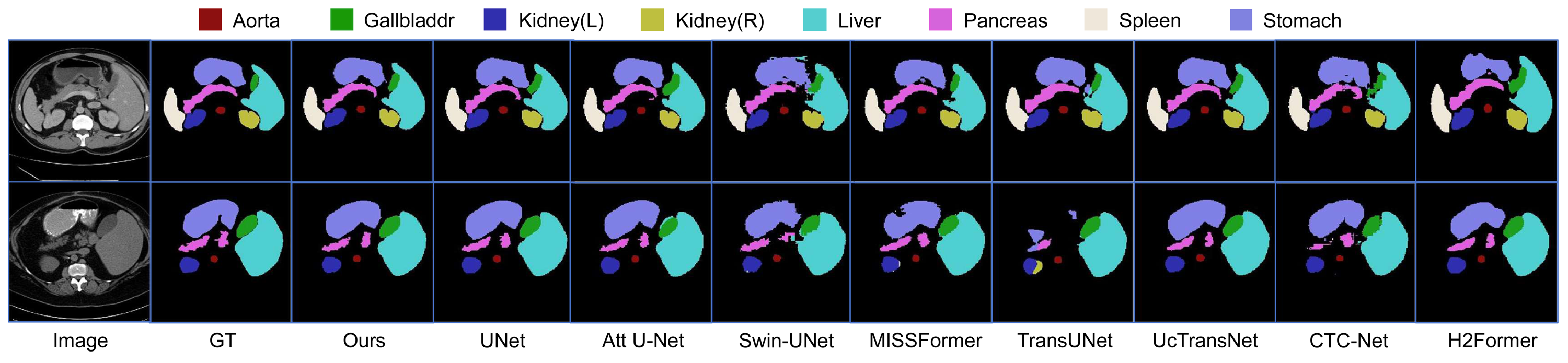}
	\caption{Visualization of segmentation results on the multi-organ Synapse dataset.} 
	\label{fig:fig3}
\end{figure}

\begin{figure}[htb]
	\centering
	\includegraphics[width=0.95\linewidth]{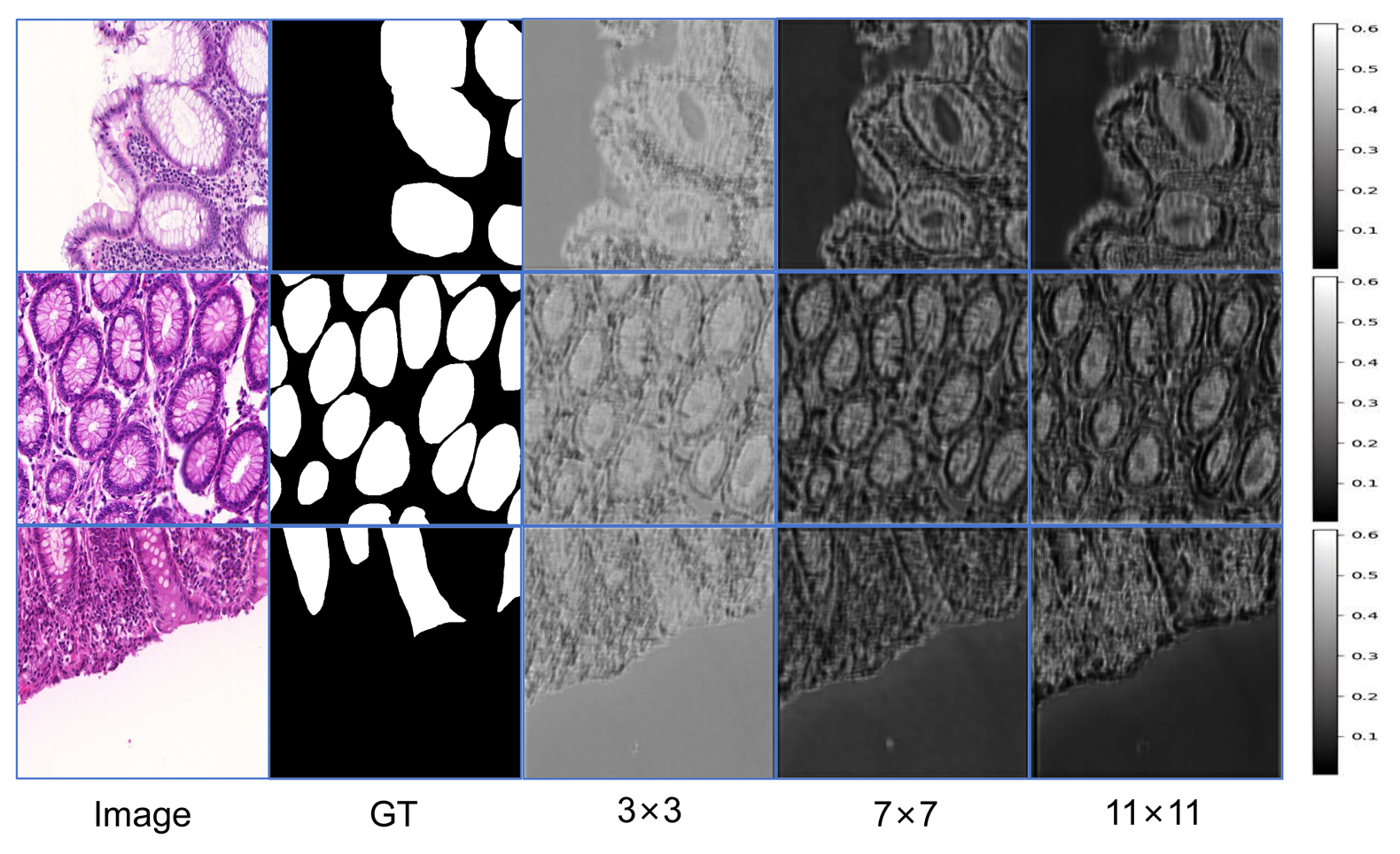}
	\caption{Visualization of the activation maps corresponding to different receptive fields.} 
	\label{fig:fig4}
\end{figure}

We visualize the segmentation results of the counterparts and our proposed method PARF-Net on the MoNuSeg, GlaS and DSB2018 datasets, as shown in Figure~\ref{fig:fig2}. PARF-Net demonstrates the better ability to capture local details, especially in distinguishing boundary of the target objects. For instance, on the GlaS dataset, the segmentation results of U-Net and Attention U-Net have obvious flaws, and the lesions are deformed by using the Transformer and hybrid counterparts, while our method correctly identifies the lesions and yields highly consistent results with the ground truth. Additionally, we also visualize the results on the multi-organ Synapse dataset, as shown in Figure~\ref{fig:fig3}. The counterparts such as TransUNet and CTC-Net are less robust to large deformation and blurred boundary of the pancreas. By comparison, our method can distinguish between different organs, and yield better segmentation results for the pancreas than the competitors.

To check if the Conv-PARF generates pixel-wise adaptive responses to different receptive fields, we visualize the activation maps $\{A_k\}_{k=1}^{K}$ as shown in Figure~\ref{fig:fig4}. With kernel size of 3$\times$3, there are no obvious differences between the activation values of lesion boundaries and neighboring pixels, and pixels within the same lesion show high homogeneity. By comparison, the activation maps corresponding to 7$\times$7 and 11$\times$11 kernels exhibit significant differences near the boundaries of the lesions, and higher variance within the same lesion. This implies different receptive fields are complementary to each other, and joint usage of them can strengthen their advantages and simultaneously compensate for their shortcomings, thus making it easier to disentangle target objects from the background. The results demonstrate the effectiveness of our proposed Conv-PARF module in extracting distinguishable features. 

\begin{table}[ht]	
	\caption{Comparisons between different model variants on the dataset GlaS.}
	\label{tab:table5}
	\begin{center}
		\begin{threeparttable}
			\begin{tabular}{c|ccccc}
				\hline
				Model variant\tnote{1} & IoU$\uparrow$ & Dice$\uparrow$ & ACC$\uparrow$ & Recall$\uparrow$ & Precision$\uparrow$\\
				\hline
				E(4,0)+D(4,0)        & 81.53±0.28 & 89.7±0.18  & 89.36±0.16  & 91.99±0.92  &87.99±0.76 \\
				E(3,1)+D(3,1)        & 81.36±0.73 & 89.58±0.45 & 89.35±0.44  & 90.83±0.89  &88.82±0.28 \\
				E(2,2)+D(2,2)        & 81.91±1    & 89.5 ±0.62 & 89.27±0.75  & 90.69±1.34  &88.79±1.69 \\
				E(1,3)+D(1,3)        & 82.22±0.45 & 90.13±0.27 & 89.85±0.27  & 92.03±0.53  &88.68±0.35 \\
				E(0,4)+D(0,4)        & 82.56±0.59 & 90.33±0.35 & 90.01±0.39  & 92.7 ±0.72  &88.45±0.66 \\
				E(1,3)+D(0,3)        & 82.77±0.42 & 90.47±0.25 & 90.29±0.25  & 91.65±0.8   &89.67±0.65 \\
				E(3,1)+D(0,1)        & 82.1±0.6   & 90.06±0.36 & 89.82±0.44  & 91.63±0.89  &88.93±1.1.17 \\
				E(2,2)+D(0,0)        & 83.02±0.74 & 90.72±0.36 & 90.61±0.28  & 91.25±0.97  &\textbf{90.52±0.68} \\
				E(2,2)+D(0,2)        & \textbf{83.76±0.65} & \textbf{91.06±0.4}   &\textbf{90.67±0.34} & \textbf{92.9±1.23}   &89.53±0.72 \\
				\hline
			\end{tabular}
			\begin{tablenotes}
				\footnotesize
				\item[1] The model variant `E(2,2)+D(0,2)' represents the proposed PARF-Net.
			\end{tablenotes}
		\end{threeparttable}
	\end{center}
\end{table}

\begin{table}[htb]	
	\caption{Comparisons between different configurations of the receptive fields in the Conv-PARF layers.}
	\label{tab:table6}
	\begin{center}
		\begin{threeparttable}
			\begin{tabular}{c|ccccc}
				\hline
				Configuration\tnote{1} & IoU$\uparrow$ & Dice$\uparrow$ & ACC$\uparrow$ & Recall$\uparrow$ & Precision$\uparrow$\\
				\hline
				(3,7,11,15,19)  & 80.9 ±0.72 & 89.28±0.45  & 88.98±0.56  & 91.36±1.52  &87.8±1.34 \\
				(3,7,11,15)     & 81.52±0.66 & 89.69±0.4   & 89.5 ±0.53  & 90.73±0.72  &89.08±1.34 \\
				(3,7)           & 83.3±0.48  & 90.79±0.3   & 90.57±0.34  & 92.23±0.91  &89.52±0.97 \\
				(3,7,11)        & \textbf{83.76±0.65} & \textbf{91.06±0.4}   &\textbf{90.67±0.34} & \textbf{92.9±1.23}   &\textbf{89.53±0.72}  \\
				\hline
			\end{tabular}
			\begin{tablenotes}
				\footnotesize
				\item[1] The configuration `(3,7,11)' is adopted in the proposed PARF-Net.
			\end{tablenotes}
		\end{threeparttable}
	\end{center}
\end{table}

\subsection{Ablation Study}
\subsubsection{Comparison between different model variants}
There are three types of layers, including static convolution, the Conv-PARF and Transformer-CNN layers, that constitute the network components of PARF-Net. Different configurations of these three types of layers result in distinct network architectures, and we make a comprehensive comparison between these model variants denoted by `E($n$,$m$)+D($p$,$q$)’, here E($n$,$m$) represents an encoder containing $n$ Conv-PARF layers and $m$ Transformer-CNN layers, and D($p$,$q$) denotes a decoder containing $p$ Conv-PARF layers and $q$ Transformer-CNN layers. We use static convolution as the default layer for all model variants, and set the number of layers to 4 for both the encoder and decode. As shown in Table~\ref{tab:table5}, the adopted architecture `E(2,2)+D(0,2)’ delivers the best results in four out of five metrics. If we introduce the Conv-PARF layers into the decoder ($p>0$), the segmentation accuracy drops significantly. For instance, with the architecture `E(2,2)+D(2,2)’, the average IoU decreases from 83.76\% to 81.91\%, and average Dice reduces from 91.06\% to 89.5\%. This validates the special role of the Conv-PARF in handling feature maps with high pixel-to-pixel differences, such as the input medical image. Additionally, utilizing static convolutions in all layers of the decoder (`E(2,2)+D(0,0)’) also yields degraded segmentation performance with mean IoU of 83.02\% and mean Dice of 90.72\%, which suggests the hybrid Transformer-CNN layers have advantages over static convolutions in reconstructing high-resolution features. We further investigate the effects of the depth of Conv-PARF layers in the encoder, and the results show the model architecture `E(1,3)+D(0,3)’ that contains one Conv-PARF layer delivers better results than other variants, indicating the Conv-PARF module should be used in lower layers. 

\subsubsection{Comparison between different configurations of the receptive fields in Conv-PARF}
In the Conv-PARF layer, the combination of different receptive fields plays an important role in capturing objects with varying shapes and scales. To figure out which configuration promises the best results, we make a comparison between different configurations of the receptive fields in Conv-PARF. As shown in Table~\ref{tab:table6}, our adopted configuration `(3,7,11)’ leads to the highest segmentation accuracy, surpassing others by a large margin. Additionally, introducing more convolutional kernels with larger receptive fields will cause evident loss of the accuracy. For instance, with configuration `(3,7,11,15,19)’, the average IoU and Dice drop to 80.9\% and 89.28\%, respectively. The reason for this may lie in loss of spatial local information and overfitting caused by large receptive fields. The comparison results demonstrate the effectiveness of the adopted configuration of receptive fields in PARF-Net.

\section{Conclusion}
Precise and robust segmentation of medical images can serve as a sound basis for computer-aided disease diagnosis. However, the objects such as lesions and organs in medical images tend to show high variance in shapes and scales, making it much difficult to accurately disentangle them from the noisy background. To resolve this problem, we propose to combine convolutions of pixel-wise adaptive receptive fields with hybrid Transformer-CNN architecture, and introduce a new method called PARF-Net for medical image segmentation. The Conv-PARF module can well adapt to inter-pixel differences in medical images, and dynamically extract spatial local information for each pixel, thus being effective in capturing objects with varying shapes and scales. We employ the Conv-PARF module in lower layers of the encoder to retain as much spatial local information as possible, making it convenient to acquire informative high-level features.

By comprehensively assessing the performance of our method on four publicly available benchmark datasets, we showcase the superior ability of PARF-Net in segmentation of both 2D and 3D medical images, and its advantages over the SOTAs. Particularly, our method surpasses the SOTAs by a large margin in pancreas segmentation on the Synapse dataset, owing to its better ability of tackling the adverse impact of large deformation and blurred boundary of the pancreas (as shown in Figure~\ref{fig:fig3}). Additionally, the ablation study demonstrates the effectiveness of the Conv-PARF and hybrid architecture, and shows their important roles in improving the segmentation accuracy. Future work will be further optimization of the model structure to achieve higher segmentation accuracy, hoping to provide help in clinical applications.  

\bibliographystyle{elsarticle-num}
\bibliography{ref}

\end{document}